\documentclass[10pt, a4paper]{article}
\usepackage[final]{lrec2026}
\usepackage[utf8]{inputenc} 
\usepackage[T1]{fontenc}    
\usepackage{hyperref}       
\usepackage{url}            
\usepackage{booktabs}       
\usepackage{array}
\usepackage{amsfonts}       
\usepackage{nicefrac}       
\usepackage{microtype}      
\usepackage{lipsum}
\usepackage{comment}
\usepackage{fancyhdr}       
\usepackage{graphicx}       
\graphicspath{{media/}}     
\usepackage{multirow} 
\usepackage{rotating}

\usepackage{geometry}
\usepackage{xcolor}
\usepackage{longtable}
\usepackage{amsmath}

\usepackage{microtype}
\usepackage{listings} % For code listings
\usepackage{subcaption}

\usepackage{authblk}

\title{Dataset Creation and Baseline Models for Sexism Detection in Hausa}

\name{Fatima Adam Muhammad\textsuperscript{1}, Shamsuddeen Muhammad Hassan\textsuperscript{2}, Isa Inuwa-Dutse\textsuperscript{3}} 
\address{Federal University Dutse\textsuperscript{1}, Imperial College London\textsuperscript{2}, University of Huddersfield\textsuperscript{3}}

\abstract{
Sexism reinforces gender inequality and social exclusion by perpetuating stereotypes, bias, and discriminatory norms. Noting how online platforms enable various forms of sexism to thrive, there is a growing need for effective sexism detection and mitigation strategies. While computational approaches to sexism detection are widespread in high-resource languages, progress remains limited in low-resource languages where limited linguistic resources and cultural differences affect how sexism is expressed and perceived. This study introduces the first Hausa sexism detection dataset, developed through community engagement, qualitative coding, and data augmentation. For cultural nuances and linguistic representation, we conducted a two-stage user study (n=66) involving native speakers to explore how sexism is defined and articulated in everyday discourse. We further experiment with both traditional machine learning classifiers and pre-trained multilingual language models and evaluating the effectiveness few-shot learning in detecting sexism in Hausa. Our findings highlight challenges in capturing cultural nuance, particularly with clarification-seeking and idiomatic expressions, and reveal a tendency for many false positives in such cases.  
\\ \newline 
\\
\textcolor{red}{Warning: This paper contains instances of sexist language to serve as examples}}

\begin{document} 
\maketitleabstract

\section{Introduction} 
\label{sec:introduction}

Sexist hate speech entails any supposition, belief, assertion, gesture or act that is aimed at expressing contempt towards a person on the basis of gender \cite{council_of_europe_2025_1680651592}. Studies on discourse analysis have shown that sexism may be expressed at different linguistic granularity levels going from lexical to discursive \cite{cameron1992feminism}. Sexism is prejudice or discrimination based on a person’s gender. It can take several forms: sexist remarks, gestures, behaviours, practices, from insults to rape or murder \cite{chiril2020annotated}. While online social media platforms such as Facebook and X have facilitated various forms of sexism to thrive, they also enable a way of retrieving relevant resources and engage with diverse participants to study sexism. 
Research on sexism detection and gender bias in language has grown rapidly in recent years. However, most existing studies focus on English and other high-resource languages, where richer linguistic resources exist  \cite{luo2025literature,rodriguez2024detecting,samory2021call}. 
Being one of the most populous language in Africa, there exist insufficient or no linguistic resources to address sexism in Hausa language. Thus, this work is aimed at developing the first corpus and baseline models for sexism detection in Hausa language. We will examine various forms of sexism and how they manifest while factoring cultural nuances and language structure. Thus, we will be addressing the following research questions (1) how to create culturally representative corpus of sexist and non-sexist texts in Hausa? and (2) how native Hausa speakers conceptualise sexism? To that effect, this study offers the following contributions: 
\begin{enumerate}
    \item We contribute the first datasets for sexism detection in Hausa language. 
    \item We conduct a two-stage user study (pilot and main) involving native Hausa speakers to explore how sexism is perceived and expressed within the community. This participatory approach ensures cultural validity and promotes community-driven resource development. 
    \item We analyse how linguistic, cultural, and societal nuances influence the manifestation and detection of sexism in Hausa. The insight is useful in providing a foundation for more inclusive and context-aware NLP systems. 
\end{enumerate}

Both data augmentation, involving translation and re-annotation of existing sexism data in English, and engagement with native speakers will be integrated in developing the detection system. The idea of taken this broad approach is to widen the data collection and not impose limitations that will affect the quantity of the data to work with. This is crucial since enriching resources through labelled data is one of the key objectives of the study. 

The remaining part of the paper is structured as follows. The next section (\S ~\ref{sec:related-work}) covers the related literature. We then present our methodology in \S~\ref{sec:methodology}, followed by the results and discussion in \S~\ref{sec:results-discussion}. The paper concludes with our findings and ethical considerations in \S~\ref{sec:conclusion}.

\section{Related Work} 
\label{sec:related-work} 

In many languages and everyday communication, sexism often manifests through explicit derogatory expressions and implicit biases, especially of women \cite{swim2004understanding,tian2025marioexist2025simple}. In high-resource languages (HLRs), substantial progress has been made in developing computational methods and resources for detecting such expressions \cite{jha2017does,samory2021call,rodriguez2024detecting,luo2025literature}. 
Relevant studies have examined the multifaceted nature of sexism and the need for robust computational systems for its detection. Using the EXIST dataset, \citet{rodriguez2024detecting} assessed its representational capacity for diverse types of sexism. Similarly, \citet{parikh2021categorizing} proposed a multi-label classification framework to capture co-occurring sexist expressions, while \citet{redondo2023anti} developed an anti-sexism alert system for identifying sexist content across multimedia platforms. \citet{bandyopadhyay2024sexism} introduced an influence-based approach for evaluating the importance of individual data points in model training for sexism detection. 
However, these advances often fail to effectively generalise to low-resource languages (LRLs) in which linguistic, cultural and pragamatic markers of sexism differ from those of HRLs. For instance, Hausa embodies distinct gendered expressions and proverbs that encode social hierarchies, moral expectations and gender norms \cite{ibrahim2025feminist}. 
Noting the scarcity of resources and the sociolinguistic diversity, there is a growing body literature exploring these challenges \cite{ekundayo2020sexisms,altininvestigating,schutz2021automatic,buscemi2025mind}. This study contributes to the literature by developing the first Hausa sexism detection dataset and evaluating the performance of multilingual and few-shot learning models.

\paragraph{Resources for Sexism Detection}  
While there is a growing body of research on sentiment analysis, offensive and hate speech detection in Hausa language \cite{muhammad2022naijasenti,adam2024detection,alkomah2022literature,zandam2023online,jahnavi2025hate,talat2016you,rizwan2020hate}, no prior study has focused specifically on sexism detection in Hausa. In contrast, several datasets have been developed for other languages to support moderation and the reduction of sexists or discriminatory content. For instance, \citet{chiril2020annotated} introduced the first French corpus annotated for sexism detection that distinguish between sexist content and narratives of sexism experienced by women. Similarly, \citet{rodriguez2020automatic} developed the \textit{MeTwo} dataset in Spanish, containing sexist expressions and attitudes on Twitter. 
For Arabic, \citet{almanea-poesio-2022-armis} presented the ArMIS corpus, annotated by individuals with varying religious backgrounds, revealing how ideological differences can affect annotation outcomes. Likewise, \citet{el2020dataset} combined automatic and manual methods to construct a dataset addressing harassment and gender-based discrimination in Arabic texts. In the Turkish context, \citet{altininvestigating} investigated the use of large language models (LLMs) for sexism detection. 

Building upon these efforts, our study introduces the first Hausa dataset for sexism detection. Our goal is to extend sexism research to one of the most populous languages in Africa.

\section{Data Collection and User Study}
\label{sec:methodology}
\begin{figure*}[ht]%[h!]
        \centering
        \includegraphics[width=0.65\linewidth]{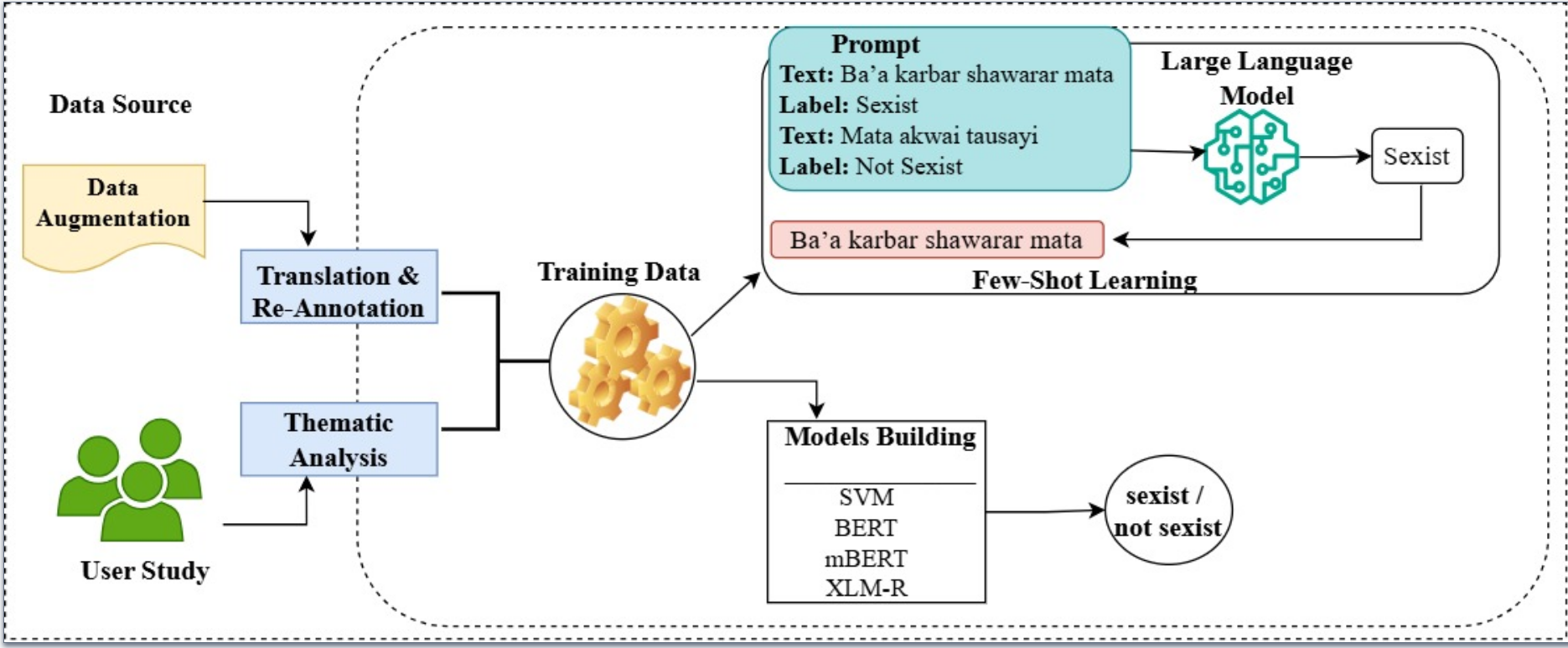}
        \caption{Overview of our approach depicting data sources (consisting of user study and data augmentation) and baseline models development.}
        \label{fig:OSD method}
    \end{figure*}

To achieve our main aim and address the research questions, we adopt a mixed-method and iterative design involving linguistic data collection, qualitative analysis, and model development (see Figure~\ref{fig:OSD method}). 

\subsection{Dataset Creation} 
The study begins with the development of a representative Hausa sexism corpus, constructed through the adaptation and extension of existing English-language sexism datasets \cite{kirk2023semeval}. To this end, we employ a combination of translation, contextual re-annotation, and native expert validation to ensure that cultural and linguistic nuances unique to Hausa are preserved. Each text is categorised as sexist or not sexist, with additional thematic tagging. 

\paragraph{User Study} 
We engage native speakers of the language in a two-stage user study to gather culturally grounded perspectives on how they perceive and articulate sexism. The pilot study (n=33) tested the clarity and relevance of survey prompts and coding categories. Participants provided open-ended definitions of sexism in Hausa, allowing for iterative refinement of the data collection and preliminary theme extraction.
Working with a different set of participants, the main user study (n = 33) builds on the pilot to include diverse Hausa speakers across regions, genders, and educational backgrounds. Respondents contributed both conceptual definitions and examples of everyday sexist expressions. The responses were analysed through qualitative thematic coding \cite{braun2006using}, yielding recurring conceptual patterns such as gender bias, role restriction, social exclusion, and moral stereotyping (Table~\ref{tab:hausa_sexism_merged_tight}). 
The user study also serves as a crowdsourced resource for model development (\S~\ref{sec:models-development}). 

%To capture a broad range of linguistic and sociocultural perspectives, 
We recruited native Hausa speakers from diverse demographic backgrounds and of varying ages (18-25 (38\%), 25-30 (38\%), 35-45 (19\%) and 45-55 (5\%)), genders (12.9\% female and 87.1\% male) and educational levels (21.9\% completed secondary school, 12.5\% have a diploma qualification, 37.5\% have a first degree and 28.1\% have a postgraduate degree). %This is to ensure representation of different perspectives on the subject.
Prior to the survey, ethical approval was obtained through the Institutional Review Board. All participants received detailed information about the study's objectives and procedures and gave informed consent before participating. All responses were anonymised to protect participant identify and confidentiality.

\paragraph{Data Augmentation} 
In addition to the user study, we employ a cross-lingual data augmentation strategy using an existing English sexism detection dataset \cite{kirk2023semeval}. 
We recruit bilingual expert translators with domain knowledge of gender and sociolinguistic contexts. We chose to employ human translators instead of relying on direct machine translation in order to prioritise semantic and cultural equivalence, ensuring that the nuances of sexist and non-sexist expressions were contextually preserved. The translated instances were re-annotated to align with local linguistic norms and gender ideologies. We leverage positive instances from \cite{adam2024detection} for the non-sexist terms in training the models; further detail and information about the data is provided in the project's github repository

\section{Sexism Detection in Hausa} 
\label{sec:results-discussion} 
We present a set of experiments to detect sexist content in Hausa messages relying both on standard feature-based and deep learning/pre-trained approaches. 

\subsection{Models} 
\label{sec:models-development}

Noting that detecting sexism is a complex, nuanced task in the language, we utilise various models including SVM \cite{cortes1995support}, BERT (and its variant mBERT) \cite{devlin2019bert}, XML-R \cite{conneau2019unsupervised} and popular LLMs for few-shot learning (GPT-5\footnote{\url{https://openai.com/gpt-5/}}, Grok\footnote{\url{https://grok.com/}} and Deepseek\footnote{\url{https://www.deepseek.com/}}).

\paragraph{Few-Shot Learning} 
For the few-shot learning FSL, we utilise relevant large language models (LLMs) fine-tuned on the examples provided by the native speakers that participated in the user study to enable the models learn to generalise from limited data. %(\S~\ref{sec:models-development}). 

    \begin{table*}[ht]
        \centering
        \small
        \begin{tabular}{lcccc}
        \hline
        \textbf{Model/Setup} &  \textbf{Accuracy (\%)} & \textbf{Precision} & \textbf{Recall} & \textbf{F1-Score} \\ 
        \hline
        GPT-5 (few-shot, 5 examples) &   \textbf{87.3} & 0.85 & \textbf{0.88} & \textbf{0.86} \\
        GPT-5 (few-shot, 10 examples) &   0.70 & 0.67 & 0.73 & 0.73 \\
        Deepseek (few-shot, 5 examples) &   0.76 & 0.85 & 0.79 & 0.82 \\
        Grok (few-shot, 5 examples) &   0.76 & 0.83 & 0.76 & 0.80 \\
        SVM   &  0.65 & 0.65 & 0.65 & 0.65 \\
        BERT  &  0.81 & 0.82 & 0.81 & 0.81 \\
        mBERT  &  0.77 & 0.77 & 0.77 & 0. 77\\
        XML-R  &  0.85 & \textbf{0.86} & 0.85 & 0.84 \\
        \hline
        \end{tabular}
        \caption{Performance comparison of trained models and few-shot sexism detection models in Hausa. Few-shot prompting (5 examples) shows better performance compared to fine-tuned baselines.} 
        \label{tab:performance}
    \end{table*} 

%The models are generally effective (recording high performance across the detection task). There is slightly better Precision than Recall, suggesting that the models are better at avoiding false alarms (labelling non-sexist content as sexist) than it is at catching every single instance of sexist content. 

\subsection{Discussion} 

In response to addressing the question of \textit{creating culturally representative corpus of sexism in Hausa}, we conducted a qualitative survey (n=66) asking participants to contribute sexist expressions in Hausa.
Similarly, to address the question of \textit{how native Hausa speakers conceptualise sexism}, we asked open questions and coded the corresponding responses to identify key thematic frames of gender-based bias, inequality, and social perception. 
In Table~\ref{tab:hausa_sexism_merged_tight}, our thematic coding revealed the following primary themes: \textit{discrimination} (43\%), \textit{inequality or bias} (26\%), \textit{stereotyping} (20\%), and \textit{prejudice/derogatory} (11\%). Across the responses, participants most frequently used the phrase \textit{wariyar jinsi} (gender discrimination) as a direct equivalent for sexism. Overall, the responses indicate that the participants conceptualise sexism primarily as a form of gender-based injustice rooted in unequal social norms. The themes in Table~\ref{tab:hausa_sexism_merged_tight} were identified inductively from the main responses using open and axial coding \cite{braun2006using} by two native speakers of Hausa, achieving high reliability (Cohen’s Kappa $k$ = 0.82; \cite{landis1977measurement}). Moreover, participants provided several representative definitions and examples of how sexism manifests linguistically and socially in Hausa. Respondents articulated sexism both conceptually and through culturally embedded expressions. The examples in Table~\ref{tab:hausa_sexism_themes} broadly fall into gender role restriction, inequality or bias, intellectual and moral stereotyping. 

    \begin{table*}[ht]
        \centering
        \setlength{\tabcolsep}{4.5pt} 
        \renewcommand{\arraystretch}{1.2} 
        \footnotesize
        \begin{tabular}{p{3.5cm}p{4.5cm}p{5.1cm}p{1.5cm}}
        \hline
        \textbf{Theme} & \textbf{Explanation (Hausa)} & \textbf{Explanation (English)} & \textbf{Freq. (\%)} \\
        \hline
        \textbf{Discrimination} & 
        \textit{Nuna wariya saboda bambancin jinsi.} & 
        Showing discrimination or unfair treatment toward someone based on gender. & 43 \\
        
        \textbf{Inequality / Bias} & 
        \textit{Nuna wariya tsakanin maza ko mata domin nuna raunin matan.} & 
        Unequal treatment of men and women, often to belittle or undermine women. & 26 \\
        
        \textbf{Stereotyping} & 
        \textit{Iyakar wani aiki ko siffa ga wani jinsi musamman mace.} & 
        Limiting certain roles or characteristics to one gender, particularly women. & 20 \\
        \textbf{Prejudice/Derogatory} & Goyon bayan rashin kyakykyawan mu'amala ko cin mutuncin mata & Promoting the mistreatment of certain women & 11 \\
        
        \hline
        \end{tabular}
        \caption{Hausa respondents’ thematic definitions of “sexism.” Themes represent major interpretive categories derived from respondents’ Hausa definitions.  
        Frequencies reflect proportions of the total responses from the main user study (n=33).}
        \label{tab:hausa_sexism_merged_tight}
    \end{table*}
        
    \begin{table*}[ht]
        \centering
        \small
        \setlength{\tabcolsep}{4.5pt} 
        \renewcommand{\arraystretch}{1.15} 
        \begin{tabular}{p{0.7cm}p{5.2cm}p{5.5cm}p{3.1cm}}
        \toprule
         & \textbf{Hausa Expression} & \textbf{English Translation} & \textbf{Theme} \\
        \midrule
        
        \multirow{4}{*}{\rotatebox[origin=c]{90}{\textbf{Definitions}}} &
        \textit{Iyakance wani aiki ko siffa ga wani jinsi musamman mace.} &
        Restricting certain roles or qualities to one gender, especially women. &
        Stereotyping \\
        & \textit{Fifita namiji akan mace.} &
        Giving preference or superiority to men over women. &
        Inequality / Bias \\
        & \textit{Sexism na nufin nuna wariya, bambanci, ko rashin adalci ga mutum saboda jinsinsa.} &
        Sexism means discrimination, inequality, or injustice toward someone based on gender. &
        Discrimination \\
        
        \midrule
        \multirow{6}{*}{\rotatebox[origin=c]{90}{\textbf{Examples}}} &
        \textit{Mace ta fi kyau a cikin gida.} &
        A woman is best suited to the home. &
        Stereotyping \\
        & \textit{Mace ba ta da hankali irin na namiji / Mace ba ta da hankali sosai.} &
        A woman is not as intelligent as a man. &
        Prejudice/Derogatory \\
        & \textit{Ba’a karbar shawarar mata.} &
        Women’s advice is not accepted. &
        Stereotype \\
        & \textit{Ba’a bawa mata amana.} &
        Women are not to be trusted. &
        Stereotype \\
        
        \bottomrule
        \end{tabular}
        \caption{The table distinguishes between abstract definitions and everyday sexist expressions, with thematic categories highlighting the sociocultural dimensions of gender bias in Hausa.}
        \label{tab:hausa_sexism_themes}
    \end{table*}

\paragraph{Detecting Sexism}
Table~\ref{tab:performance} shows a summary of the results from the trained models. The models are generally effective, recording high performance across the detection task. However, the models struggle to understand subtle and nuance language as exemplified in Table~\ref{tab:qualitative-examples}. For instance, clarification-seeking or sarcastic expressions (as in sentences ending with 'ne') are often misclassified. There is slightly better Precision than Recall, suggesting that the models are better at avoiding false alarms (labelling non-sexist content as sexist) than it is at catching every single instance of sexist content.

%########################%Qualitative Examples Table (Interpretability)
    \begin{table*}[ht]
        \centering
        \setlength{\tabcolsep}{4.5pt} 
        \renewcommand{\arraystretch}{1.2} 
        \footnotesize
        %\centering
        %\small
        \begin{tabular}{p{4.5cm}ccc p{4cm}}
        \hline
        \textbf{Text (Hausa)} & \textbf{Human Label} & \textbf{Model Prediction} & \textbf{Correct?} & \textbf{Remark} \\
        \hline
        \textit{Mata adon gari} & Sexist & Sexist & \checkmark & Implicitly reinforces gender as ornamental. \\
        \textit{Babu bambanci tsakanin mata da maza} & Not Sexist & Not Sexist & \checkmark & Correctly recognises inclusive phrasing. \\
        \textit{Mace ko za tafi kyau a cikin gida ne} & Not Sexist & Sexist & $\times$ & Seeking clarification. \\
        \textit{Mace mai kamar maza} & Sexist & Sexist & \checkmark & Correctly flags mocking tone. \\
        \textit{Mace ta san zafin nema ne} & Sexist & Not Sexist & $\times$ & Clarification-seeking/Sarcasm \\ %Questionable capability.?
        \hline
        \end{tabular}
        \caption{Qualitative examples of Hausa sexism detection. The model, GPT-5 (few-shot, 5 examples), performs well on explicit and implicit cases but occasionally misses culturally nuanced expressions.}
        \label{tab:qualitative-examples}
    \end{table*}

%\paragraph{Detecting Sexism}
%Table~\ref{tab:performance} shows a summary of the results from the trained models. The models are generally effective, recording high performance across the detection task. However, the models struggle to understand subtle and nuance language as exemplified in Table~\ref{tab:qualitative-examples}. For instance, clarification-seeking or sarcastic expressions (as in sentences ending with 'ne') are often misclassified. There is slightly better Precision than Recall, suggesting that the models are better at avoiding false alarms (labelling non-sexist content as sexist) than it is at catching every single instance of sexist content. 

\section{Conclusion} 
\label{sec:conclusion}
Noting the challenges associated with sexism detection in low-resource languages, this study contributes the first systematic effort toward detecting sexism in Hausa language. Our approach include the construction of a culturally representative dataset through engagement with native speakers and data augmentation strategy based on existing sexism data in English language. 
We experiment with both traditional and pre-trained multilingual models to demonstrate how sexism detection can be extended to low-resource languages. The results highlight both the challenges of cross-lingual transfer and the promise of few-shot learning for scalable detection task. This work underscores the importance of integrating community knowledge and cultural nuance into NLP resource development, especially in low-resource settings. Future research will explore context-aware modelling and the broader application of this approach to other LRLs to promote inclusive and socially responsible language technologies.

\subsection{Ethical Considerations and Limitations}
We understand that this study presents some ethical considerations and potential harms that may arise. Firstly, the models may fail to recognise subtle or context-specific forms of sexism due to data scarcity and language dynamism. We incorporated input from the local community via user study to ensure that the dataset used for training is as inclusive as possible. We acknowledged the following limitations: 
\begin{itemize}
    \item[-] The small size of available datasets may lead to suboptimum performance, especially in detecting nuanced or subtle forms of sexism. Moreover, the user study data is skewed because only a handful of the participants are females (12.9\%) and future work will need to ensure higher participation. 
    \item[-] Sexism is context-dependent, and the meaning of a phrase or statement may change based on social, cultural, and situational factors. While effective at detecting patterns of sexism in text, the model may struggle to capture this contextual nuance. For example, clarification-seeking or sarcastic expressions may be misclassified as shown in Table~\ref{tab:qualitative-examples}, and the model may not distinguish between genuine intent and socially accepted behaviours that inadvertently perpetuate gender stereotypes. %The data is also limited in this regard. 
\end{itemize}

\section{Bibliographical References}
\label{sec:reference}

\bibliographystyle{lrec2026-natbib}
\bibliography{main}

\end{document}